\documentclass[conference]{IEEEtran}
\IEEEoverridecommandlockouts

\usepackage{cite}
\usepackage{amsmath,amssymb,amsfonts}
\usepackage{graphicx}
\usepackage{textcomp}
\usepackage{xcolor}
\usepackage[abs]{overpic}
\usepackage{algorithm}
\usepackage[end]{algpseudocode}
\usepackage{etoolbox}
\usepackage{subcaption}
\usepackage{hyperref}
\usepackage[inline]{enumitem}

\def\BibTeX{{\rm B\kern-.05em{\sc i\kern-.025em b}\kern-.08em
    T\kern-.1667em\lower.7ex\hbox{E}\kern-.125emX}}
\begin{document}

\title{
Towards a Safe Real-Time Motion Planning Framework for Autonomous Driving Systems: A Model Predictive Path Integral Approach \\
}

\author{
\IEEEauthorblockN{Mehdi Testouri}
\IEEEauthorblockA{\textit{University of Luxembourg} \\
\textit{Interdisciplinary Centre for, } \\
\textit{Security, Reliability and Trust (SnT)} \\
mehdi.testouri@uni.lu}
\and
\IEEEauthorblockN{Gamal Elghazaly}
\IEEEauthorblockA{\textit{University of Luxembourg} \\
\textit{Interdisciplinary Centre for, } \\
\textit{Security, Reliability and Trust (SnT)} \\
gamal.elghazaly@uni.lu}
\and
\IEEEauthorblockN{Raphael Frank}
\IEEEauthorblockA{\textit{University of Luxembourg} \\
\textit{Interdisciplinary Centre for, } \\
\textit{Security, Reliability and Trust (SnT)} \\
raphael.frank@uni.lu}
}

\maketitle

\begin{abstract}

Planning safe trajectories in Autonomous Driving Systems (ADS) is a complex problem to solve in real-time. The main challenge to solve this problem arises from the various conditions and constraints 
imposed by road geometry, semantics and traffic rules, as well as the presence of dynamic agents. Recently, Model Predictive Path Integral (MPPI) has shown to be an effective framework for optimal motion planning and control in robot navigation in unstructured and highly uncertain environments.
In this paper, we formulate the motion planning problem in ADS as a nonlinear stochastic dynamic optimization problem that can be solved using an MPPI strategy. 
The main technical contribution of this work is a method to handle obstacles within the MPPI formulation safely. In this method, obstacles are approximated by circles that can be easily integrated into the MPPI cost formulation while considering safety margins. 
The proposed MPPI framework has been efficiently implemented in our autonomous vehicle and experimentally validated using three different primitive scenarios. 
Experimental results show that generated trajectories are safe, feasible and perfectly achieve the planning objective. 
The video results as well as the open-source implementation are available at 
\textcolor{magenta}{\href{https://github.com/sntubix/mppi}{https://github.com/sntubix/mppi}}.

\end{abstract}
\vspace{5px}

\section{Introduction}

Motion planning is a complex and challenging problem in robotics and autonomous systems \cite{paden2016survey}. A motion planner has to consider system dynamics, constraints, sensor uncertainties, and real-time constraints to ensure that the system safely and efficiently navigates a given environment.
The ultimate goal of a motion planner is to generate a feasible, collision-free, comfortable, and fail-safe trajectory, while strictly satisfying a set of spatio-temporal constraints. Typically these constraints allow generated trajectories to (a) avoid collision with obstacles surrounding the vehicle, (b) ensure dynamic and kinematic feasibility as well as vehicle actuation capabilities (e.g. limits of steering angles, steering rate, maximum acceleration and braking torques), and (c) take into account the complex geometry and topology of the road environment. Violating any of these constraints may lead to unpredictable and risky behaviour of the autonomous vehicle. 

Motion planning for robots and autonomous systems somehow amounts to solving a complex optimization problem. The main challenge to solve a motion planning problem derives mainly from the non-convexity of the underlying optimization problem. Furthermore, the spatio-temporal constraints of dynamic obstacles hinder most of the motion planning approaches to reach a feasible trajectory within predefined real-time constraints \cite{manzinger2020using}. 
Ensuring these constraints are respected is fundamental in safety-critical situations, where the autonomous vehicle must be able to anticipate in a timely manner to prevent a potential collision.

\begin{figure}[!t]
    \centering
    \includegraphics[scale=0.32]{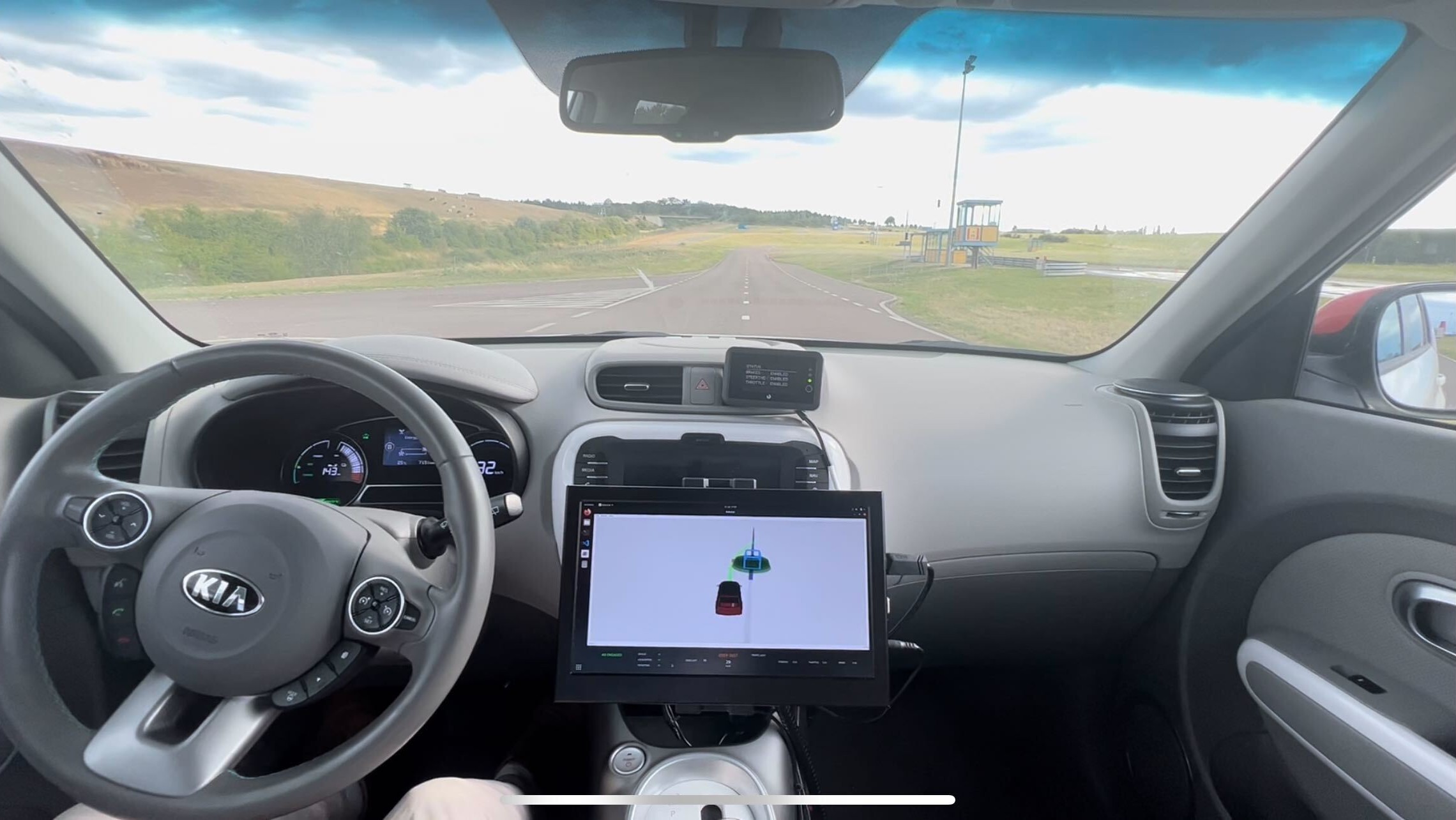}
    \caption{\small Validation of the MPPI-based motion planner on a closed track using a modified KIA Soul EV equipped for autonomous driving. The figure depicts the object avoidance scenario whereby a virtual object was placed on the path.}
    \label{fig:track}
\end{figure}

Motion planning is quite an established research topic in the vast state-of-the-art of autonomous vehicles \cite{paden2016survey, gonzalez2015review}. Several approaches have been developed in the literature to solve the above-described trajectory planning challenge. Recently there has been an increasing interest in using learning-based methods, especially reinforcement learning \cite{zhang2021safe, rezaee2021motion}. Nevertheless, solutions based on these methods are unlikely to meet the safety requirements of autonomous vehicles and it becomes difficult to verify and interpret, thus putting their reliability in question, especially for safety-critical situations \cite{xu2022trustworthy}. Numerical optimization has also been a widely accepted approach to solve motion planning problems \cite{zhang2020optimization, ziegler2014trajectory}. 
Although efficient numerical optimization algorithms exist to converge 
to a solution, one limitation of these algorithms is that they may become stuck in a local minimum due to non-convexity. Decoupling longitudinal and lateral motions \cite{manzinger2020using}
and estimating the driving corridor \cite{ding2019safe} is one way around reformulating 
the problem as a convex one. This approximation may, however, result in slightly infeasible trajectories.

Methods from control theory have gained significant interest in motion planning. 
These methods formulate the motion problem as an optimal control problem subject to constraints on both the states and inputs of the systems \cite{paden2016survey}. 
Model Predictive Control (MPC) is one of the efficient methods to solve constrained optimal control problems using the receding horizon concept. 
MPC iteratively solves an optimal control sequence along a finite-time horizon while minimising the cost function of a planning objective \cite{yoon2009model}. A major limitation of MPC in motion planning lies in the high computational cost it needs, especially with high-dimensional nonlinear systems and the difficulty to consider obstacles in its formulation. Recently, Model Predictive Path Integral (MPPI) has shown to be an effective framework for optimal motion planning and control for robots and autonomous systems \cite{mohamed2023towards, mohamed2022autonomous}. The key idea of MPPI is to transform the optimal control problem into an expectation over the integral of all possible trajectories \cite{williams2017model, williams2018information, williams2016aggressive, williams2018robust}. 
This enables to solve the optimal control problem using Monte Carlo simulations by forward sampling a large number of trajectories that can be evaluated efficiently using parallel processing, e.g. using GPUs. 
The way most MPPI frameworks consider avoiding obstacles in their formulation is still in its infancy with obstacles considered as cost maps \cite{buyval2019model}. Furthermore, most of these frameworks are tailored for robot navigation in unstructured and highly uncertain environments \cite{mohamed2020model, mohamed2023towards, mohamed2021sampling}. In the context of Autonomous Driving Systems (ADS), only very few results are available \cite{buyval2019model, guardini2022minimal}. 
This paper takes a step forward by developing an MPPI framework that systematically considers obstacles and safety in an autonomous driving setting. 
Obstacles provided by the perception system are approximated by a finite set of circles while considering a safety margin. This way of representing obstacles makes it possible to penalize trajectories at risk of collision more easily. 
The proposed MPPI framework is efficiently implemented in our autonomous vehicle and experimentally validated in three different primitive scenarios. The first scenario is to allow the vehicle to merge a lane starting from a lane neighbourhood while the second is to perform a turning manoeuvre to avoid an obstacle. The third scenario is to have the autonomous vehicle follows an obstacle in the same lane. Although these scenarios are individually simple, they indeed meet the requirements of generated trajectories for complex scenarios if combined together.
Experimental results show that generated trajectories are safe, feasible and perfectly achieve the planning objective.

The rest of this paper is organized as follows. Section \ref{sec:problem} presents some mathematical preliminaries, the modelling of the vehicle as well as the formulation of the motion planning as a constrained stochastic nonlinear dynamic optimization problem.
The proposed MPPI motion planning framework is presented in Section \ref{sec:mppi-framework}. 
The experimental results are presented in Section \ref{sec:results}. 
Finally, we draw conclusions and discuss future works in Section \ref{sec:conclusion}.

\section{Preliminaries and Problem Statement}
\label{sec:problem}

The objective of this paper is to plan feasible safe trajectories for autonomous vehicles to navigate their environment. This section presents a preliminary mathematical background for vehicle modelling and the foundations of MPPI used to develop the approach of this work.

\subsection{Problem Formulation}

\begin{figure}[!t]
\centering
\begin{overpic}[scale=0.95, trim={1.45cm 1.9cm 1.0cm 2.0cm},clip]{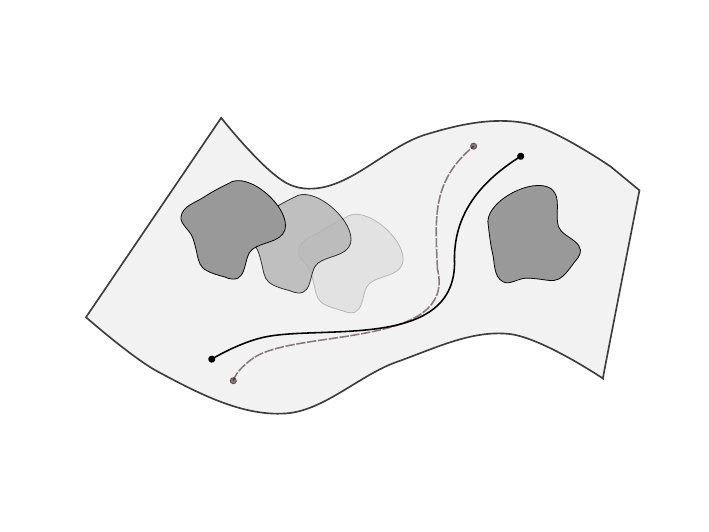}
\put(10,47){\footnotesize $\mathcal{X} \subset \mathbb{R}^n$}
\put(55,85){\footnotesize $\mathcal{O}_{j}^{d}(t_0)$}
\put(92,80){\footnotesize $\mathcal{O}_{j}^{d}(t_1)$}
\put(119,72){\footnotesize $\mathcal{O}_{j}^{d}(t_2)$}
\put(198,80){\footnotesize $\mathcal{O}_{i}^{s}$}
\put(60,10){$\mathbf{x}_{0}$}
\put(182,129){$\mathbf{x}_{f}$}
\put(43,23){$\mathbf{x}_{0}^r$}
\put(205,120){$\mathbf{x}_{f}^r$}
\end{overpic}
\caption{\small Definition of the motion planning problem. Given an initial state of the vehicle $\mathbf{x}_0$, a reference state $\mathbf{x}_t^f$, a set of static and $\mathcal{O}_{i}^{s}$ and $\mathcal{O}_{j}^{d}(t)$. The objective is to generate a collision-free sequence of state $\mathbf{x}_t \in \left [ t_0, \, t_f \right ]$ while maintaining state feasibility}
\label{fig:planning-problem}
\end{figure}

The motion planning problem in autonomous driving systems can be defined in terms of four main components $\mathcal{P}: \left< \, \mathcal{V}, \, \mathcal{S}, \, \mathcal{B}, \, \mathcal{J} \, \right>$, namely a vehicle motion model $\mathcal{V}$, a scene representation model $\mathcal{S}$, a boundaries model $\mathcal{B}$, and an objective $\mathcal{J}$. Let us consider the following stochastic nonlinear discrete-time dynamical system to represent the motion model of the vehicle

\begin{subequations}
\begin{align}
   \mathcal{V}: \,\,\, & \mathbf{x}_{t+1} = \mathcal{F} \left( \mathbf{x}_{t},\mathbf{u}_{t} + \epsilon_t \right) + \mathbf{w}_{t}, 
   \label{eq:state-model} \\
   \,\,\, &  \text{s.t.} \,\,\, \mathbf{h} \left( \mathbf{x}_{t},\mathbf{u}_{t} \right) \leq 0
\end{align}
\label{eq:vehicle-state-model}
\end{subequations}

\noindent with $\mathbf{x}_t \subset \mathcal{X} \in \mathbb{R}^{n}$ is the system state, $\mathbf{u}_t \subset \mathcal{U} \in \mathbb{R}^{m}$ is the control input, $\epsilon_t \thicksim \mathcal{N} \left(0, \Sigma \right)$ is an additive zero-mean Gaussian noise to control inputs with covariance $\Sigma$, $\mathbf{w}_{t} \in \mathbb{R}^{n}$ represents unknown but bounded state disturbances, and $\mathcal{F}:\, \mathcal{X} \times \, \mathcal{U} \rightarrow \mathcal{X}$ is a state transition function. The function $\mathbf{h} \left( \mathbf{x}_{t},\mathbf{u}_{t} \right)$ encapsulates all constraints imposed by the system itself, e.g. state and control lower and upper limits as well as kinematic 
constraints. 
The scene representation $\mathcal{S}$ defines a model of obstacles to be avoided by the vehicle whether these obstacles are static or dynamic. Let $\mathcal{O}_{i}^{s} \subset \mathcal{X}, \, \forall i \in \left \{ 1,\, 2\, \dots, \, n_s \right \}$ be a set of static obstacles in vehicle's environment and $\mathcal{O}_{j}^{d} \left ( t \right ) \subset \mathcal{X}, \, \forall j \in \left \{ 1,\, 2\, \dots, \, n_d \right \}$. Dynamic obstacles impose additional constraints on the free space available for planning the motion of the vehicle as depicted in Fig. \ref{fig:planning-problem}. These additional constraints account for collision-free conditions along a given planning horizon $T = t_f - t_0$, where $t_0$ and $t_f$ are the start and end times of the planning cycle, respectively. The subsets of the state space occupied by obstacles must be avoided while planning vehicle motion. One can represent the scene model as 
\begin{subequations}
\begin{align}
   \mathcal{S}: \, & \mathbf{x}_{t} \notin \cup_{i=1}^{n_s} \mathcal{O}_{i}^{s}, \\ 
   & \mathbf{x}_{t} \notin \cup_{j=1}^{n_d} \mathcal{O}_{j}^{d} \left ( t \right ), \,\, \forall t \in \left [ \, t_0, \, t_f \, \right ] 
\end{align}
\label{eq:scene-model}
\end{subequations}

Unlike robot motion planning in an unstructured environment, 
autonomous vehicles have to respect the constraints posed by the geometry of a highly structured road environment, often represented by a high-definition map \cite{elghazaly2023high}. Basically, this condition can be fulfilled if the vehicle state trajectory is allowed to follow a given reference state $\mathbf{x}_{t}^{r} \in \left [\, t_0, \, t_f\, \right ]$ along the planning horizon. This objective can be satisfied by adding a function $\phi \left ( \mathbf{x}_{t}, \mathbf{x}_{t}^r \right )$ to the cost function to be minimized

\begin{equation}
   \mathcal{B}: \, \phi \left ( \mathbf{x}_{t}, \mathbf{x}_{t}^r \right ) \in \mathbb{R}^{+} \,\, \forall t \in \left [ \, t_0, \, t_f \, \right ] 
\label{eq:road-model}
\end{equation}

A motion planner seeks to generate a feasible, smooth and collision-free trajectory. More precisely, the planner must handle collisions with surrounding obstacles in the vehicle's environment, safely as defined by \eqref{eq:scene-model}. Furthermore, it must ensure dynamic and kinematic feasibility as well as vehicle actuation capabilities (e.g. limits of steering angles, steering rate, maximum acceleration and braking torques) as defined in \eqref{eq:vehicle-state-model}. Finally, it must take into account the complex geometry and topology of the road environment defined in \eqref{eq:road-model}.
The motion planning problem $\mathcal{P}$ can be expressed mathematically by combining \eqref{eq:vehicle-state-model}, \eqref{eq:scene-model}, and \eqref{eq:road-model} into the following optimization problem
\begin{align}
\min _{\mathbf{u}}  \, \, \, &  \mathbb{E}\left[\pi\left(\mathbf{x}_{f}, \, \mathbf{x}_{f}^r \right)+\sum_{t=t_0}^{t_f-1}\left(\phi \left(\mathbf{x}_{t}, \mathbf{x}_{t}^r\, \right)+\frac{1}{2} \mathbf{u}_{t}^\top R \mathbf{u}_{t}\right)\right]\!, \nonumber \\
\text {s.t.} 
\quad & \mathbf{x}_{t+1}= \mathcal{F}\left(\mathbf{x}_{t}, \mathbf{u}_{t}+\epsilon_{t}\right), \,\,\, \epsilon_{t} \sim \mathcal{N}(\mathbf{0}, \Sigma), \nonumber \\
& \mathbf{h}(\mathbf{x}_t, \mathbf{u}_t) \leq 0, \nonumber \\
& \mathbf{x}_{t} \notin \cup_{i=1}^{n_s} \mathcal{O}_{i}^{s}, \,\, \mathbf{x}_{t} \notin \cup_{j=1}^{n_d} \mathcal{O}_{j}^{d} \left ( t \right ), \nonumber \\
& \mathbf{x}_t, \, \mathbf{x}_t^r \in \mathcal{X}, \,\,\,  \mathbf{u}_{t} \in \mathcal{U}, 
\label{eq:optimization}
\end{align}

\subsection{Vehicle Motion Model}

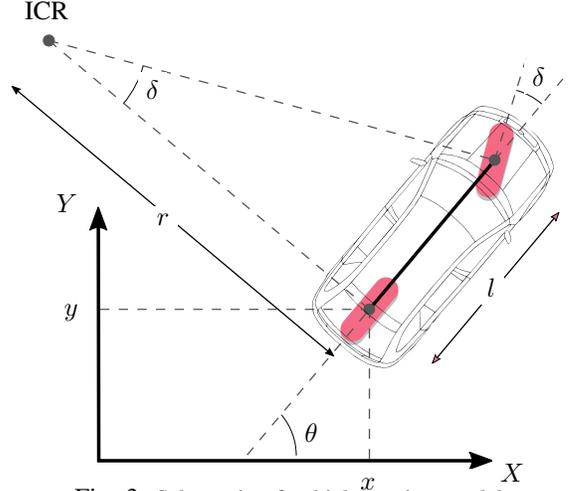
\begin{figure}[!t]
\centering
\begin{overpic}[scale=0.65, trim={0.4cm 0.4cm 0.4cm 0.3cm},clip]{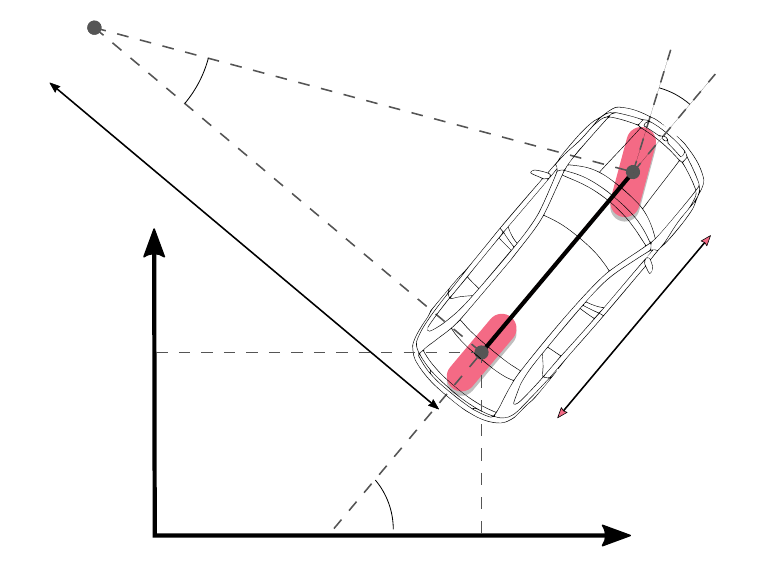}
\put(137,-8){\colorbox{white}{$x$}}
\put(25,58){\colorbox{white}{$y$}}
\put(116,10){\colorbox{white}{$\theta$}}
\put(202,145){\colorbox{white}{$\delta$}}
\put(56,140){\colorbox{white}{$\delta$}}
\put(10,170){\colorbox{white}{ICR}}
\put(60,92){\colorbox{white}{$r$}}
\put(185,65){\colorbox{white}{$l$}}
\put(190,-5){\colorbox{white}{$X$}}
\put(22,97){\colorbox{white}{$Y$}}
\end{overpic}
\caption{\small Schematic of vehicle motion model}
\label{fig:kinematic}
\end{figure}

For simplicity and to reduce computational complexity, this work adopts a bicycle model to represent the motion model of the vehicle as shown in Fig. \ref{fig:kinematic}. Let $x$ and $y$ be the coordinates of a vehicle position with respect to a given reference map coordinate system $\mathcal{R}_m: \left \{O,\, X,\, Y\right \}$, $\theta$ be the orientation of the vehicle with respect to $\mathcal{R}_m$ and $\delta$ is steering angle. Let $r$ be the radius of the circle centred at the instantaneous centre of rotation (ICR), and the linear velocity of the vehicle $v$ is related to angular velocity about ICR as $v = \dot{\theta} r$. From the triangle formed by the vehicle and the ICR, we have $\tan{\left(\delta\right)} = l/r$, where $l$ is the wheelbase of the vehicle. Thus, one can express the motion model of the vehicle as
\begin{equation}
\begin{split}\frac{d}{dt}\begin{pmatrix} x\\ y\\ \theta\\ v \\ \delta \end{pmatrix} 
=\begin{pmatrix} v\cos(\theta)\\ v\sin(\theta) \\ v\tan(\delta)/l \\ a \\ \omega \end{pmatrix}\end{split}
\label{eq:bicycle}
\end{equation}
\noindent where $a$ is the vehicle linear acceleration and $\omega$ is the steering rate. 
One can easily rewrite the vehicle motion model given by \eqref{eq:bicycle} in the form of \eqref{eq:state-model} with the state vector $\mathbf{x} = \left( x,\, y,\, \theta,\, v,\, \delta \right )^T$ and control input $\mathbf{u} = \left( a,\, \omega \right )^T$. 

\section{MPPI Motion Planning Framework}
\label{sec:mppi-framework}

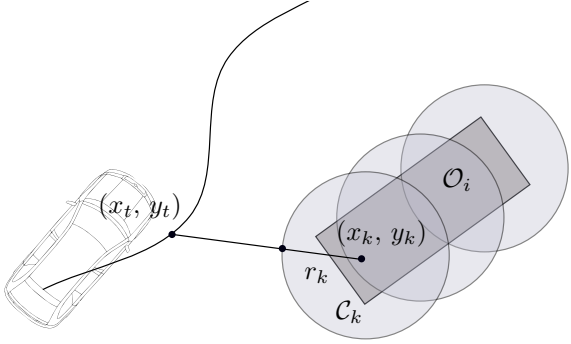
\begin{figure}[!t]
\centering
\begin{overpic}[scale=0.6, trim={0.0cm 1.0cm 0.0cm 1.0cm},clip]{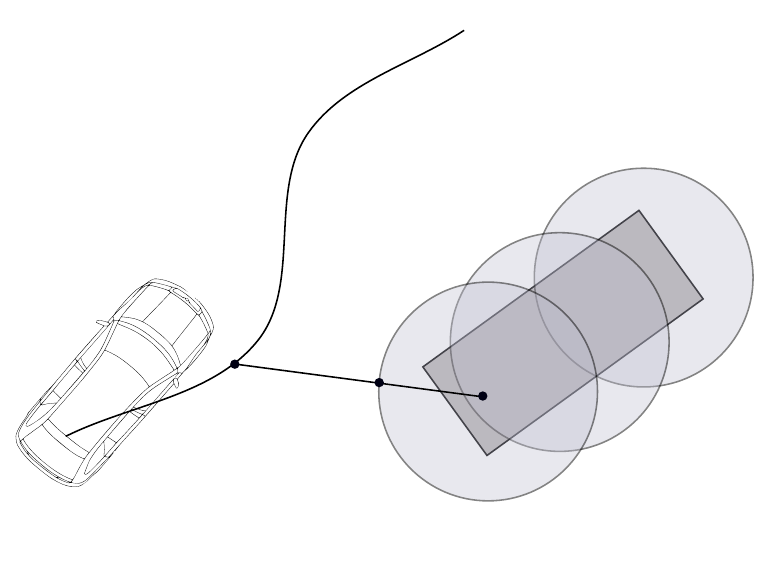}
\put(170,60){$\mathcal{O}_{i}$}
\put(130,10){$\mathcal{C}_{k}$} 
\put(118,25){$r_k$}
\put(40,50){$(x_{t},\, y_t)$}
\put(130,40){$(x_{k},\, y_k)$}
\end{overpic}
\caption{\small A perception obstacle $\mathcal{O}_{i}$ approximated by circles $\mathcal{C}_{k} = \left \{ x_{k}, y_{k}, r_{k}\, \right \}, \forall k = 1, \, 2, \, 3$. } 
\label{fig:obstacles}
\end{figure}

The complex optimization problem in (\ref{eq:optimization}) is challenging to solve mainly because of the non-convexity of the search space. Furthermore, the spatio-temporal constraints of dynamic obstacles hinder most of the motion planning approaches to reach a feasible trajectory to meet real-time constraints. Ensuring these constraints are respected is fundamental in safety-critical situations, where the autonomous vehicle must be able to anticipate in a timely manner to prevent a potential collision. 
MPPI offers an efficient way to solve (\ref{eq:optimization}). MPPI can be used to solve optimization problems subject to nonlinear dynamics and non-convex constraints. The main idea of MPPI is to randomly sample a huge number of $M$ trajectories generated by Monte Carlo simulations of the vehicle state model \eqref{eq:state-model} in real-time. These trajectories are evaluated in \textit{parallel} by computing the cost-to-go from $t_0$ to $t_f$ along each trajectory. 
For a given sample period $\Delta t$, the horizon $t \in \left [t_0, \,t_f \right ]$ is discretized into $T$ samples. Let $S(\tau_i)$ denotes the cost of trajectory $\tau_i, \, \forall i = 1,\, 2,\, \dots, \, M$. 

\begin{equation}
    S(\tau_i) = \pi\left(\mathbf{x}_{T}, \, \mathbf{x}_{T}^r \right)+\sum_{t=0}^{T-1} \tilde{q} \left(\mathbf{x}_{t},\, \mathbf{x}_{t}^r, \, \mathbf{u}_{t},\, \epsilon_{t} \right),
\end{equation}

\noindent where $\pi\left(\mathbf{x}_{T}, \, \mathbf{x}_{T}^r \right)$ is referred to the terminal cost with the objective to force the final trajectory state $\mathbf{x}_{T}$ to be as close as to a selected target point $\mathbf{x}_{T}^r$. The term $\tilde{q} \left(\mathbf{x}_{t},\, \mathbf{x}_{t}^r, \, \mathbf{u}_{t},\, \epsilon_{t} \right)$ is referred to as the running cost with the goal to satisfy the constraints of the planning problem as defined in \eqref{eq:optimization}. For the sake of simplicity, one can decouple the running cost into two components
\begin{equation}
    \tilde{q} \left(\mathbf{x}_{t},\, \mathbf{x}_{t}^r, \, \mathbf{u}_{t},\, \epsilon_{t} \right) = \tilde{q}_{x} \left(\mathbf{x}_{t},\, \mathbf{x}_{t}^r \right) +
    \tilde{q}_{u} \left( \mathbf{u}_{t},\, \epsilon_{t} \right)
\end{equation}

\noindent where $\tilde{q}_{x} \left(\mathbf{x}_{t},\, \mathbf{x}_{t}^r \right)$ considers all costs that depend on the system state $\mathbf{x}_{t}$ and reference state $\mathbf{x}_{t}^r$ trajectories, and $\tilde{q}_{u} \left( \mathbf{u}_{t},\, \epsilon_{t} \right)$ is typically quadratic cost to optimize the input sequence
\begin{equation}
    \tilde{q}_{u} \left( \mathbf{u}_{t},\, \epsilon_{t} \right) = \alpha \, \mathbf{\epsilon}_{t}^\top R \mathbf{u}_{t} +
    \mathbf{u}_{t}^\top R \mathbf{\epsilon}_{t} +
    \frac{1}{2} \mathbf{u}_{t}^\top R \mathbf{u}_{t}
\end{equation}

\noindent with $R \in \mathbb{R}^{m \times m}$ is a symmetric 
positive semi-definite input weight matrix and $\alpha$ is a parameter that is used to prioritize noise against optimizing inputs, thus controlling the way the state space is explored. It is typically defined as
\begin{equation}
    \alpha = \frac{\gamma-1}{2\gamma}, \, \, \gamma \in \mathbb{R}^{+}
\end{equation}

\noindent with $\gamma$ being a positive hyperparameter which is given as an input to the MPPI algorithm. 
On the other hand, the state-dependent cost $\tilde{q}_{x} \left(\mathbf{x}_{t},\, \mathbf{x}_{t}^r \right)$ encapsulate all costs needed to force the MPPI to respect the constraints of obstacles and follow the reference trajectory as stated in \eqref{eq:scene-model} and \eqref{eq:road-model}. For the generated trajectory $\mathbf{x}_{t}$ to be as close as possible to the reference state trajectory $\mathbf{x}_{t}^r$, a weighted quadratic cost of the error between $\mathbf{x}_{t}$ and $\mathbf{x}_{t}^r$ is considered. 
We may then express $\tilde{q}_{x}$ as
\begin{equation}
    \tilde{q}_{x} = \frac{1}{2} \left ( \mathbf{x}_{t} - \mathbf{x}_{t}^r\right)^T H \left ( \mathbf{x}_{t} - \mathbf{x}_{t}^r\right) + \tilde{q}_{o} 
\end{equation}

\noindent where $H \in \mathbb{R}^{n \times n}$ is a symmetric positive definite weight matrix and $\tilde{q}_{o}$ is the cost to avoid obstacles. Our methodology to consider obstacles in MPPI is to approximate the occupied area by each obstacle $\mathcal{O}_{i}$ by a finite set of circles $\mathcal{C}_{k} = \left \{ x_{k}, y_{k}, r_{k}\, \right \}$ as illustrated in Fig. \ref{fig:obstacles}, where $x_{k}, y_{k}$ are the coordinates of centre and $r_k$ is the radius. The cost $\tilde{q}_{o}$ is considered such that distance between $(x_{k}, y_{k})$ and $(x_{i}, y_{i})$ is greater than the sum the radius $r_k$ of a safety margin $d_k$. 
This safety margin is used to anticipate the velocity of the vehicle. 
More details are given in Section \ref{sec:results}.

\subsection{Algorithm}

Algorithm \ref{algo:mppi} sketches a pseudocode of the MPPI motion planner. The algorithm starts by reading the vehicle state and the list of perception objects from the localization and perception modules of the ADS, respectively. 
Perception objects are then approximated by circles. Our MPPI implementation checks the generated noise to guarantee that a set of constraints are never exceeded, namely : $\omega_{max}$, $a_{max}$, $a_{min}$, and the target speed $v_G$. The constraints provide safety and feasibility to the trajectory. Given the cost-to-go $S(\tau_i)$ and an initial input sequence $\mathbf{U} = \left[\mathbf{u}_{0}, \mathbf{u}_{1}, \dots,\mathbf{u}_{T-1}\right]^{\top}$, one can update the input sequence iteratively using \cite{williams2017model, mohamed2020model} taking into account all individual costs contributing to the cost-to-go the cost-to-go $S(\tau_i)$ 
\begin{equation}
 \mathbf{u}_{t} \leftarrow \mathbf{u}_{t}+\frac{\sum_{m=0}^{M-1} \exp \bigl( \frac{-1}{\lambda} \left[\tilde{S}\left(\tau_{t, m}\right) -\tilde{S}_{\min} \right] \bigr) \mathbf{\epsilon}_{t, m}}{\sum_{m=0}^{M-1} \exp \bigl(\frac{-1}{\lambda} \left[\tilde{S}\left(\tau_{t, m}\right) -\tilde{S}_{\min} \right]\bigr)}
\label{eq:update-input}
\end{equation}

\noindent where $\tilde{S}_{\min}$ is mainly used to guarantee the numerical stability of the algorithm while keeping the optimality conditions of the algorithm. Practically the term $\tilde{S}_{\min}$ is considered as the 
cost of the trajectory with the minimum cost-to-go value among the set of sampled trajectories \cite{mohamed2023towards, mohamed2022autonomous}.
A final step in the MPPI algorithm is to smooth the resulting input 
sequence using a Savitzky–Golay filter \cite{savitzky1964smoothing} with coefficients $Z$ in a similar way as \cite{mohamed2023towards, mohamed2022autonomous}. 
Unlike methods that use MPPI to control the dynamic of a system directly 
by applying the first input in the optimal sequence to the system, for trajectory planning, we concatenate the optimal input sequence together with their corresponding optimal state to constitute the generated trajectory $\{\mathbf{x}_t^{\star}, \mathbf{u}_t^{\star}\}$, which subsequently is fed to the control module of the ADS.

\begin{algorithm}[ht!]
\caption{Real-Time MPPI Trajectory Generation}
\label{algo:mppi}
\hspace*{\algorithmicindent} \textbf{Given:} \\
\hspace*{1cm} $M$: Size of trajectory rollouts \\
\hspace*{1cm} $T$: Horizon timesteps \\
\hspace*{1cm} $ \mathbf{U}$: Initial input sequence \\
\hspace*{1cm} $ \mathcal{F}$: Vehicle dynamic model \\
\hspace*{1cm} $ \mathbf{x}_t^r$: Reference state \\
\hspace*{1cm} $\pi, \phi$: Cost functions \\
\hspace*{1cm} $\lambda, \gamma, \Sigma, R$: hyper-parameters \\
\hspace*{1cm} $D$: safety distance parameters \\
\hspace*{1cm} $Z$: Savitzky–Golay filter parameters 
\begin{algorithmic}[1] 
\While {not done}
    \State $\mathbf{x}_{0} \leftarrow$ GetInitialState( ), $\mathbf{x}_{0} \in \mathbb{R}^{n}$
    \State $\mathcal{O}_{i}^s, \mathcal{O}_{j}^d \leftarrow$ GetObstacles( ), $\mathcal{O}_{i}^{s}, \mathcal{O}_{i}^{d} \subset \mathcal{X}, \, \forall i, \,j$
    \State $ \mathbf{C}_{k} \leftarrow$ ApproximateObstacles($\mathcal{O}_{i}^{s}, \mathcal{O}_{i}^{d}$), $\mathbf{C}_{k} \in \mathcal{C}, \, \forall k$, 
    \State $ \mathbf{\epsilon} \leftarrow$ GenerateRandomNoise($\Sigma$), $\mathbf{\epsilon} \in \mathbb{R}^{M \times T}$
    \State $\tilde{S}\left(\tau_{m}\right) \leftarrow $ GetInitialTrajectoryCost(), $\tilde{S}\left(\tau_{m}\right) \in \mathbb{R}^{M}$
    \For{$m \leftarrow 0$ \textbf{to} $M-1$ \textit{in parallel}}
        \State $\mathbf{x} \leftarrow \mathbf{x}_{0}$
        \For{$t \leftarrow 0$ \textbf{to} $T-1$}
            \State $\mathbf{x}_{t+1} \leftarrow \mathcal{F}\left(\mathbf{x}_{t}, \mathbf{u}_{t}+ \epsilon_{t}\right)$
            \State $\tilde{q} \leftarrow \tilde{q} \left(\mathbf{x}_{t},\, \mathbf{x}_{t}^r, \, \mathbf{u}_{t},\, \epsilon_{t}, \, \mathbf{C}_{k}, \, D, \, R, \, \gamma \right)$ 
            \State $\tilde{S}\left(\tau_{t+1, m}\right) \leftarrow \tilde{S}\left(\tau_{t, m}\right)+\tilde{q}$
        \EndFor
        \State $\tilde{S}\left(\tau_{m}\right) \leftarrow \tilde{S}\left(\tau_{t+1, m}\right)+ \phi\left(\mathbf{x}_{T}\right)$
    \EndFor
    \State $\tilde{S}_{\min} \leftarrow \min _{m}[\tilde{S}\left(\tau_{m}\right)]$
    \For{$t \leftarrow 0$ \textbf{to} $T-1$}
        \State $\mathbf{u}_{t} \leftarrow \mathbf{u}_{t}+\frac{\sum_{m=0}^{M-1} \exp \bigl( \frac{-1}{\lambda} \left[\tilde{S}\left(\tau_{t, m}\right) -\tilde{S}_{\min} \right] \bigr) \mathbf{\epsilon}_{t, M}}{\sum_{m=0}^{M-1} \exp \bigl(\frac{-1}{\lambda} \left[\tilde{S}\left(\tau_{t, m}\right) -\tilde{S}_{\min} \right]\bigr)}$
    \EndFor 
    \State $\mathbf{u}^{\star} \leftarrow \text{SavitzkyGolayFilter}(Z, \, \mathbf{u})$
    \For{$t \leftarrow 0$ \textbf{to} $T - 1$}
        \State $\mathbf{x}_{t+1}^{\star} \leftarrow \mathcal{F}\left(\mathbf{x}_{t}^{\star}, \mathbf{u}_{t}^{\star}\right)$
    \EndFor
    \For{$t \leftarrow 0$ \textbf{to} $T - 2$}
        \State $\mathbf{u}_{t} = \mathbf{u}_{t+1}^{\star}$
    \EndFor
    \State \textbf{if completed} return optimal trajectory $\{\mathbf{x}_t^{\star}, \mathbf{u}_t^{\star}\}$
\EndWhile
\end{algorithmic}
\end{algorithm}
\section{Experimental Results}
\label{sec:results}

The MPPI-based motion planner has been implemented and tested on an existing autonomous driving platform \cite{9062824}. Three driving scenarios were tested on a real track using virtual objects, namely a lane merge, an object avoidance and a vehicle following. Those scenarios were selected as they form primitive functions to a more elaborate motion planner.

\subsection{Implementation Details}
The implementation has been done in C++ and integrated into an existing in-house autonomous driving stack. The MPPI being a method relying heavily on parallel computing, the implementation uses OpenCL, which provides the flexibility of using any compatible GPU. In the context of this paper, the MPPI was run on an Intel integrated GPU hinting an efficient resource usage by relieving the main GPU for other tasks that could be involved in autonomous driving.

\subsection{Driving Scenarios}
Three driving scenarios were assessed. Those scenarios were deemed relevant as they represent building blocks of what could be a higher-level motion planner.

\subsubsection{Lane Merge}
This scenario assesses the ability of the vehicle to switch lanes smoothly, a recurrent situation in everyday driving. The vehicle starts from a lane that is parallel to the target lane and has to properly merge into it. 

\subsubsection{Object Avoidance}
The goal of this scenario is to assess the ability of the motion planner to generate a trajectory that avoids an object or an obstacle obstructing the current lane. A typical real-world occurrence would be a construction site or a stopped vehicle.

\subsubsection{Vehicle Following}
This scenario is similar to object avoidance with the added constraint that the vehicle cannot leave the current lane thus forcing it to adjust its speed to keep a safe distance or even stop if the obstacle is not moving.

\subsection{Cost Formulation}
The cost function is broken down into five parts each of them pertaining to a component of the autonomous driving task. Let $P_t = (x_t,y_t)$ be the position of the vehicle and $\theta_{t}$ and $v_t$ be respectively the yaw and speed of the vehicle at timestep $t$. Let $W_t = (x_{W_t},y_{W_t})$ be the closest waypoint and $\theta^{r}_t$ and $v^{r}_t$ be the reference yaw and speed. The $c_{dist}$ cost measures the distance between the vehicle and $W_t$ thus penalising diverging trajectories.
\begin{equation}
    c_{dist} = ||(P_t - W_t)||^2
\end{equation}

Let $T_P$ be the target point, the $c_{target}$ cost penalises trajectories that do not get closer to the target point, thus preventing the vehicle from going backwards.
\begin{equation}
c_{target}=
\left\{
\begin{aligned}
  &1 \ if \ ||(P_t - T_P)|| > ||(P_{t-1} - T_P)|| \\
  &0 \ otherwise
\end{aligned}
\right.
\end{equation}

In order to ensure a smooth lane merge, the vehicle needs to stay sufficiently parallel to the target lane while closing in. The $c_{yaw}$ cost measures the difference between the current yaw and the reference yaw, both wrapped in $[-\pi, \pi]$.
\begin{equation}
    c_{yaw} = |\theta_t - \theta^{r}_t|^2
\end{equation}

This cost also proved to be helpful in following a curved path. The $c_{speed}$ cost measures the difference between the current speed and the reference speed.
\begin{equation}
    c_{speed} = |v_t - v^{r}_t|^2
\end{equation}

MPPI objects are represented by circles of centers $O^{k}_t$ and radius $r^{k}_t$. Let $d_{obj}$ be the minimal object distance and $d_{safe}$ the desired safe distance:
\begin{equation}
\left\{
\begin{aligned}
  &d_{obj} = \min_{\forall k}(||P_t - O^{k}_t|| - r^{k}_t)\\
  &d_{safe} = d_{safe_c} v_t + d_{safe_0}
\end{aligned}
\right.
\end{equation}
where $d_{safe_c}$ is the safe distance coefficient and $d_{safe_0}$ is the minimal safe distance. The $c_{safe\_dist}$ cost is formulated as follow:
\begin{equation}
    c_{safe\_dist} = \max(d_{safe} - d_{obj}, 0)^2
\end{equation}

In the case of the object avoidance scenario, 
the cost is only considered if there is a collision \textit{i.e} $d_{obj} \leq 0$, otherwise $c_{safe\_dist}$ is set to 0. The total step cost is a weighted sum of all sub costs.
\begin{equation}
\begin{aligned} 
  c_{total} = \ &15.0 c_{min\_dist} + 7.0c_{target} + 120.0c_{yaw}\\
                   &+ 5.0c_{speed} + 25.0c_{safe\_dist}
\end{aligned}
\end{equation}

\subsection{Experimental Setup}

\begin{table}[]
    \centering
    \begin{tabular}{p{1.5cm} p{1.5cm} p{1.5cm}}
    \hline
    \hline
    Parameter & Value & Unit \\
    \hline
    $\lambda$ & 150 & \\
    $\mathrm{M}$ & 2560 & \\
    $\mathrm{T}$ & 16 & \\
    $\Delta t$ & 0.25 & $\mathrm{s}$\\
    $\sigma_{\omega}$ & 0.05 & $\mathrm{rad/s}$\\
    $\sigma_{a}$ & 0.85 & $\mathrm{m/s^2}$\\
    $\omega_{max}$ & 0.11 & $\mathrm{rad/s}$\\
    $a_{max}$ & 1.1 & $\mathrm{m/s^2}$\\
    $a_{min}$ & -2.5 & $\mathrm{m/s^2}$\\
    $d_{safe_c}$ & 1.36 & \\
    $d_{safe_0}$ & 11 & $\mathrm{m}$\\
    $v_G$ & 30 & $\mathrm{km/h}$\\
    \hline
    \hline
\end{tabular}
    \caption{MPPI parameters}
    \label{tab:parameter}
\end{table}

A total of $M = 2560$ trajectories were generated in parallel. The trajectory horizon is set to 4 seconds divided into $T = 16$ steps of $\Delta t = 0.25$ seconds. $\lambda$ was set to 150 while the noise variance was $\Sigma = \mathrm{Diag}(\sigma_{\omega}, \sigma_{a}) = \mathrm{Diag}(0.05, 0.85)$. Theses parameters were obtained throught experiments. The maximum steering rate was $\omega_{max} = 0.11\mathrm{rad/s}$ while the maximum and minimum accelerations were $a_{max} = 1.1\mathrm{m/s^{2}}$ and $a_{min} = -2.5\mathrm{m/s^{2}}$. The constraints on $\omega$ and $a$ help to ensure a safe and feasible trajectory. A target speed $v_G = 30\mathrm{km/h}$ was set along the path. The safe distance coefficient was set to $d_{safe_c} = 1.36$ and the minimum safe distance to $d_{safe_0} = 11$. This corresponds approximately to a safe distance of 18 meters at 30km/h accounting for the length of the vehicle. The parameters are summarized on Table \ref{tab:parameter}. A Savitzky-Golay filter was also applied on the generated input $\mathbf{u}_{t}$ and is shown in (\ref{eq:sg}) where $\mathbf{u}^{*}_{t}$ is the filtered input.

\begin{equation}\label{eq:sg}
    {\mathbf{u}^{*}_{t}={\frac {1}{35}}(-3\mathbf{u}_{t-2}+12\mathbf{u}_{t-1}+17\mathbf{u}_{t}+12\mathbf{u}_{t+1}-3\mathbf{u}_{t+2})}
\end{equation}

The MPPI was run at 20Hz on an Intel i7-8700T and its integrated UHD Graphics 630. The tests were conducted on a real track using a modified KIA Soul EV equipped for autonomous driving as shown on Figure \ref{fig:track}. The objects required by the scenarios were obtained by the simulator module of the ADS running the MPPI. This module provides programmable virtual objects that can be tailored for specific use cases. The use of virtual objects was motivated by the need of a safe and reliable mean to assess the motion planner. The objects dimensions were set to the ones of an average car for both the object avoidance and object following scenarios. In the case of the object avoidance, a relatively tight margin of 0.7m was prescribed between the vehicle and the object while overtaking.

\subsection{Results}

\begin{figure}[!h]
\centering
\begin{subfigure}{0.5\textwidth}
\centering
    \includegraphics[scale=0.24, trim={0.0cm 0.0cm 0.0cm 0.0cm},clip]{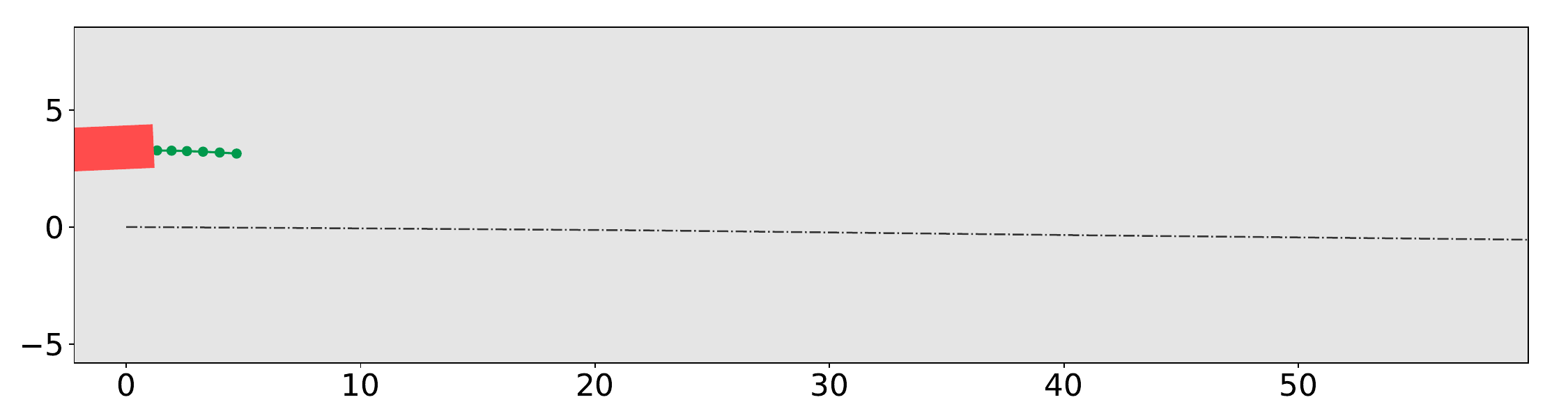}
    \caption*{$t\approx2$s}
\end{subfigure}
\hfill
\begin{subfigure}{0.5\textwidth}
\centering
    \includegraphics[scale=0.24, trim={0.0cm 0.0cm 0.0cm 0.0cm},clip]{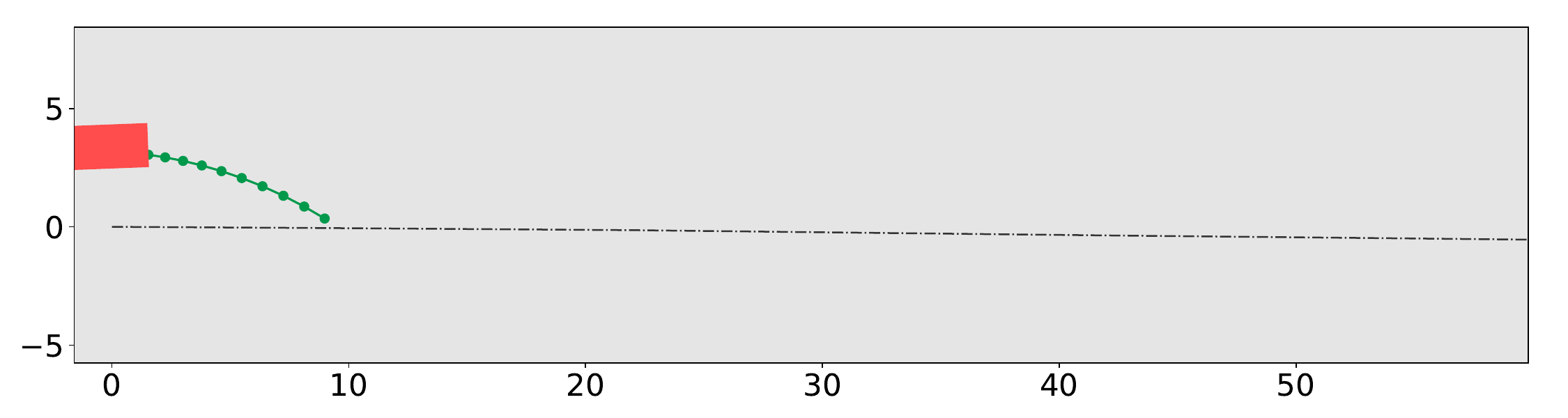}
    \caption*{$t\approx3$s}
\end{subfigure}
\hfill
\begin{subfigure}{0.5\textwidth}
\centering
    \includegraphics[scale=0.24, trim={0.0cm 0.0cm 0.0cm 0.0cm},clip]{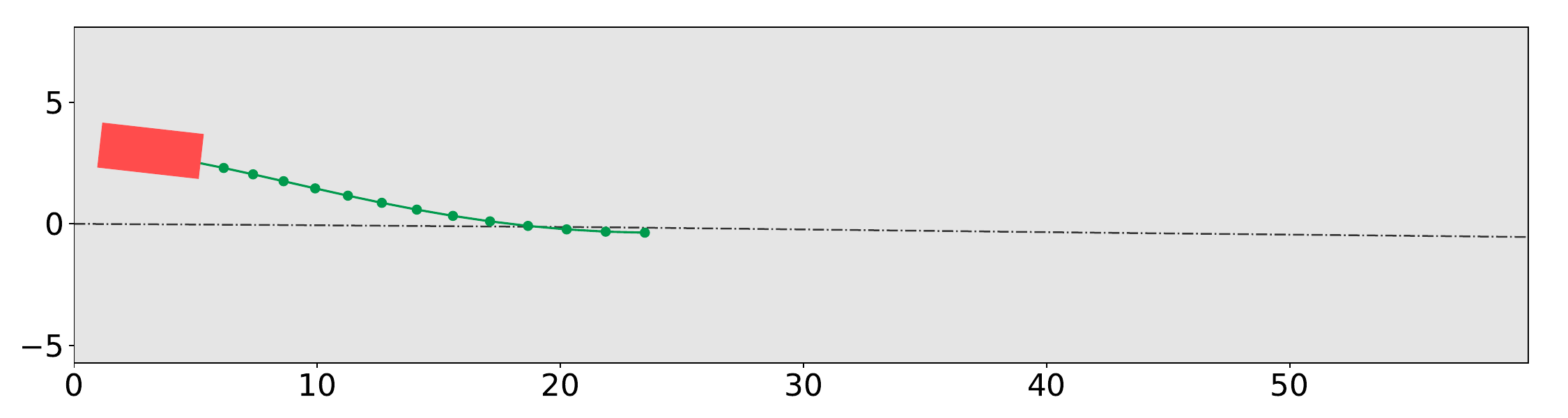}
    \caption*{$t\approx5$s}
\end{subfigure}
\hfill
\begin{subfigure}{0.5\textwidth}
\centering
    \includegraphics[scale=0.24, trim={0.0cm 0.0cm 0.0cm 0.0cm},clip]{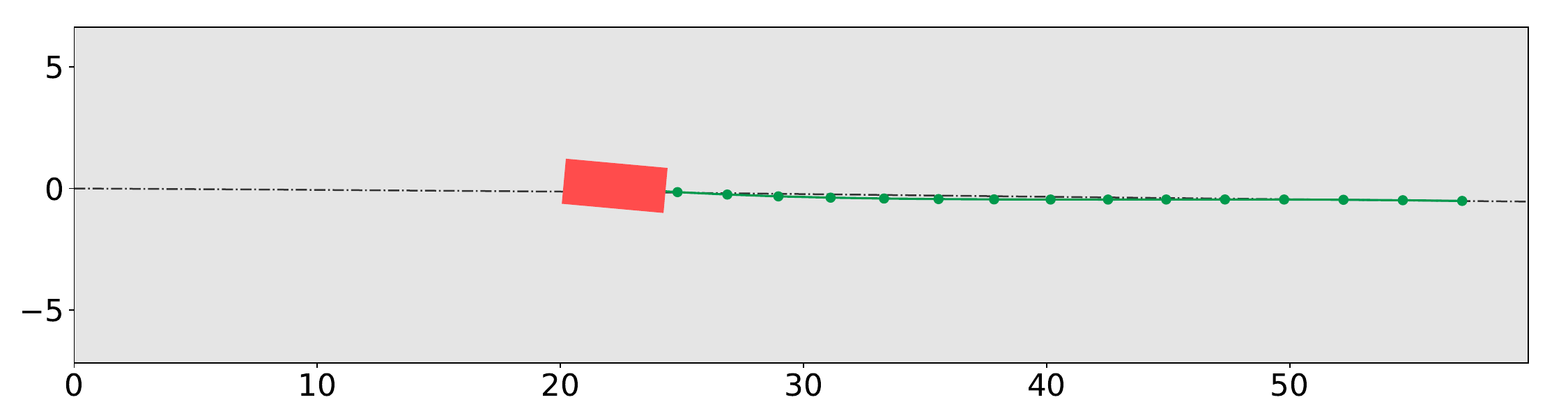}
    \caption*{$t\approx8$s}
\end{subfigure}
\hfill
\begin{subfigure}{0.5\textwidth}
\centering
    \includegraphics[scale=0.21, trim={3.5cm 0.0cm 0.0cm 0.0cm},clip]{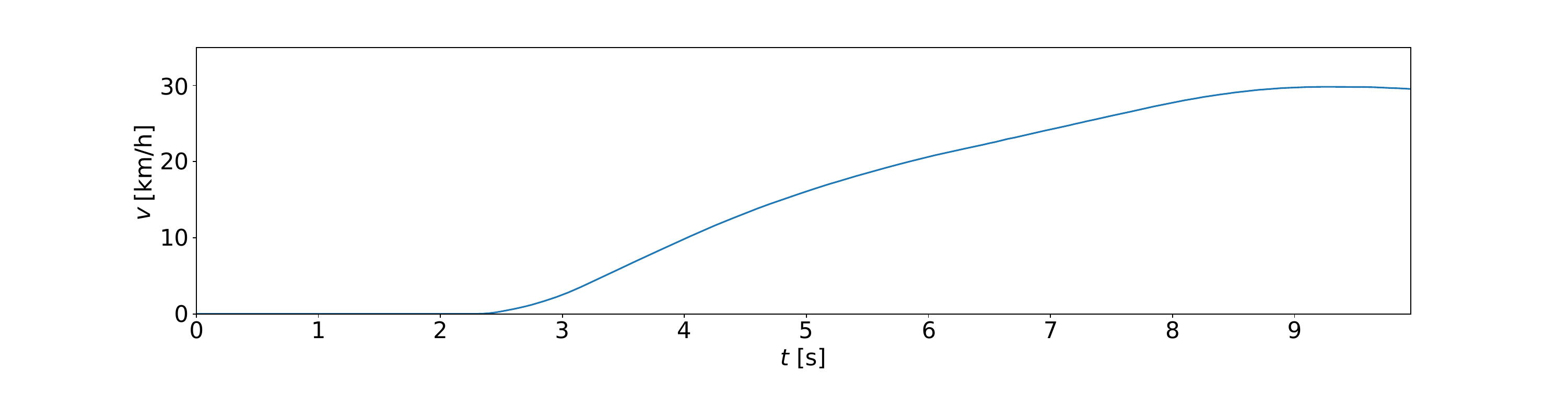}
    \caption*{Speed}
\end{subfigure}
\hfill
\begin{subfigure}{0.5\textwidth}
\centering
    \includegraphics[scale=0.21, trim={3.5cm 0.0cm 0.0cm 0.0cm},clip]{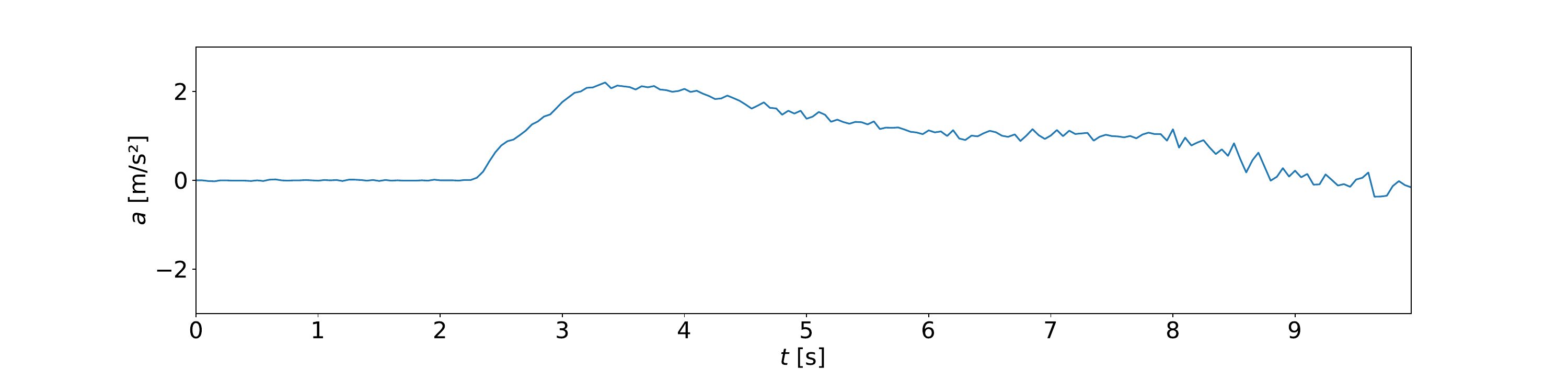}
    \caption*{Acceleration}
\end{subfigure}
\hfill
\begin{subfigure}{0.5\textwidth}
\centering
    \includegraphics[scale=0.21, trim={3.5cm 0.0cm 0.0cm 0.0cm},clip]{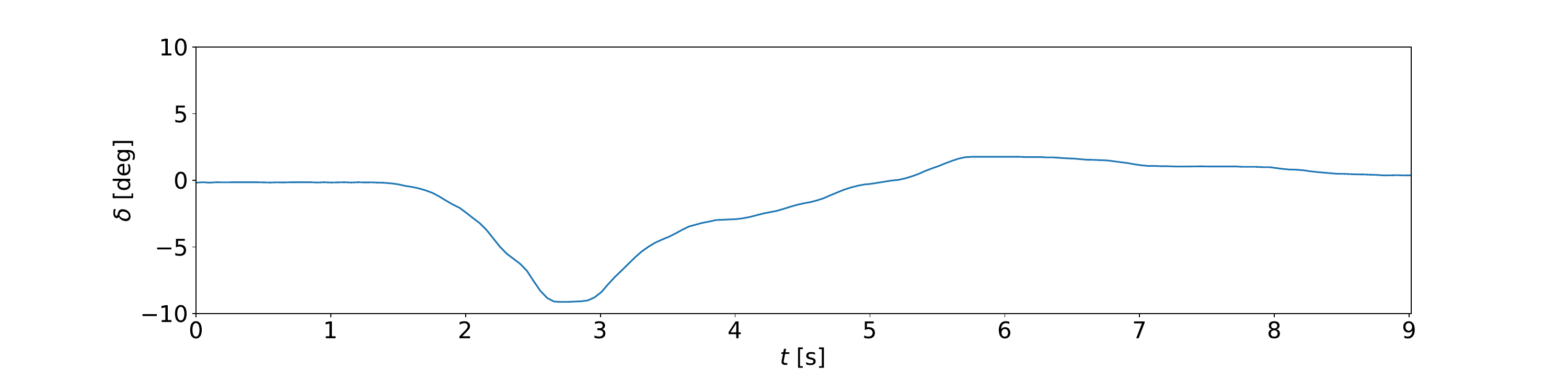}
    \caption*{Steering}
\end{subfigure}
\hfill
\caption{Lane merge scenario.}
\label{fig:merge}
\end{figure}

\begin{figure}[!h]
\centering
\begin{subfigure}{0.5\textwidth}
\centering
    \includegraphics[scale=0.24, trim={0.0cm 0.0cm 0.0cm 0.0cm},clip]{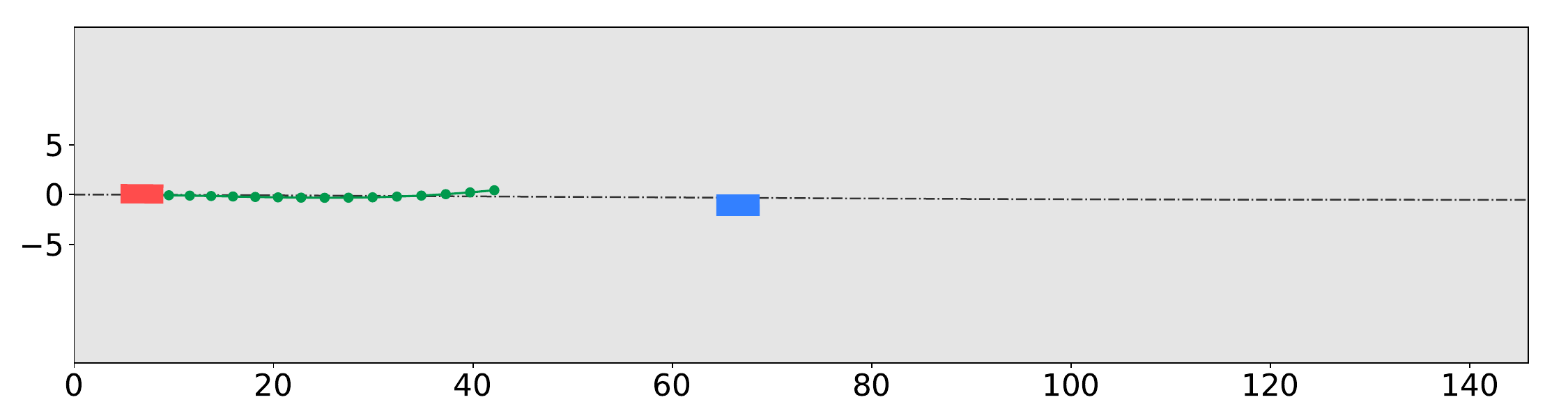}
     \caption*{$t\approx1$s}
\end{subfigure}
\hfill
\begin{subfigure}{0.5\textwidth}
\centering
    \includegraphics[scale=0.24, trim={0.0cm 0.0cm 0.0cm 0.0cm},clip]{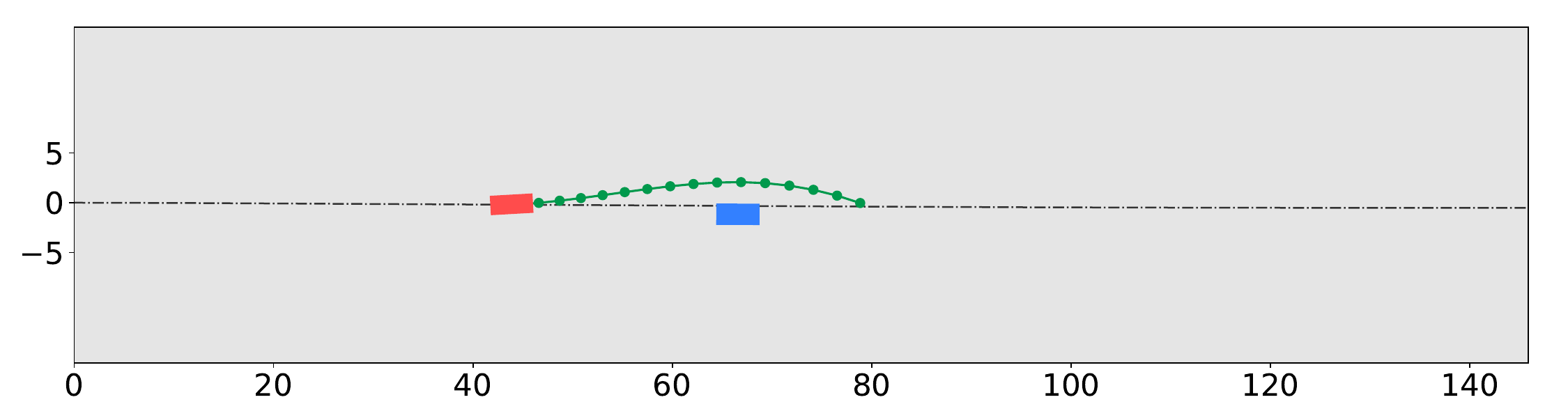}
     \caption*{$t\approx6$s}
\end{subfigure}
\hfill
\begin{subfigure}{0.5\textwidth}
\centering
    \includegraphics[scale=0.24, trim={0.0cm 0.0cm 0.0cm 0.0cm},clip]{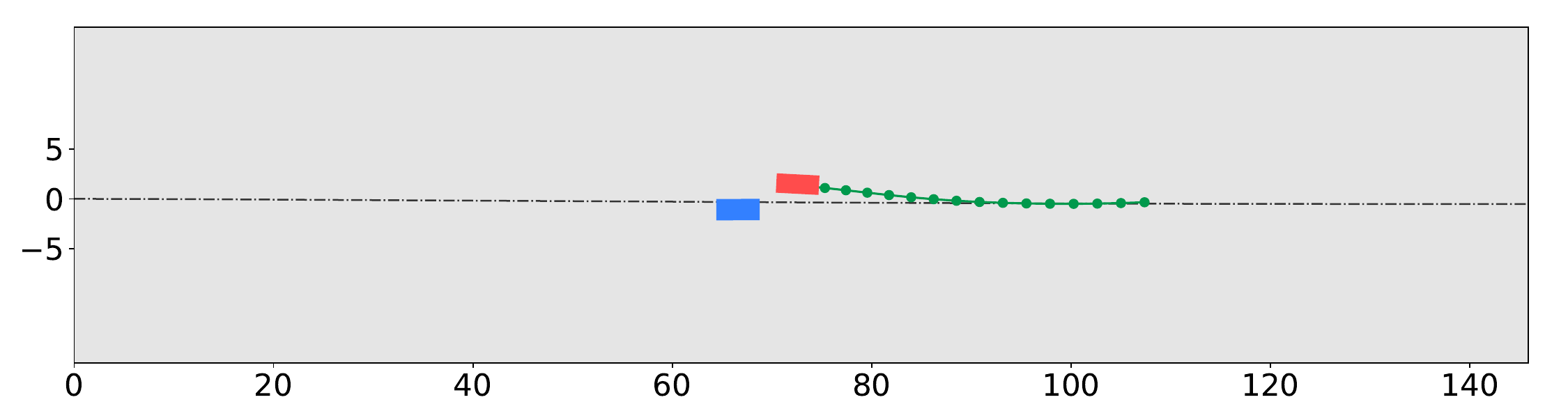}
     \caption*{$t\approx9$s}
\end{subfigure}
\hfill
\begin{subfigure}{0.5\textwidth}
\centering
    \includegraphics[scale=0.24, trim={0.0cm 0.0cm 0.0cm 0.0cm},clip]{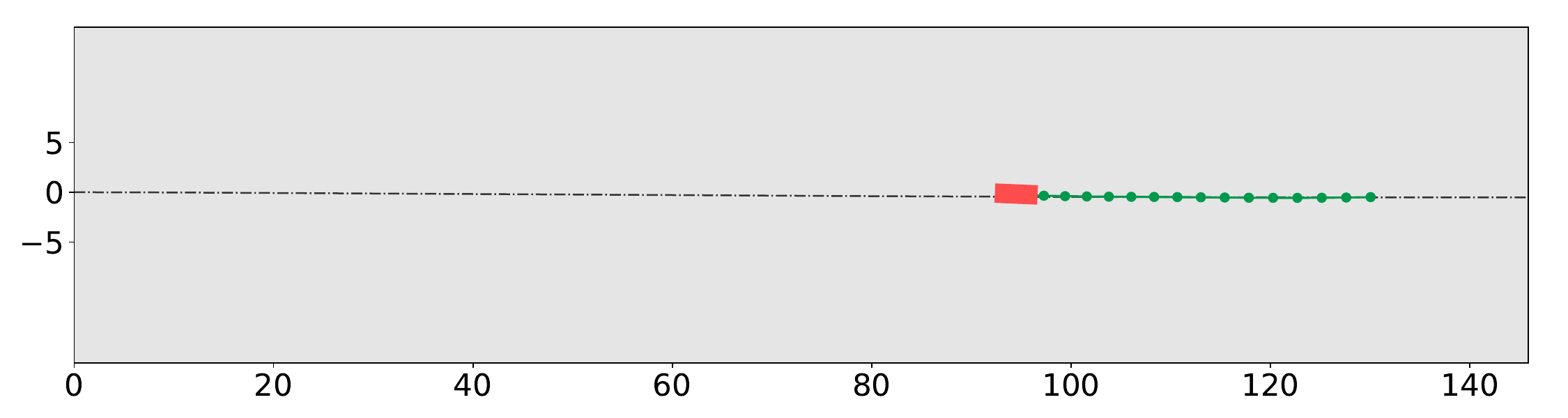}
     \caption*{$t\approx12$s}
\end{subfigure}
\hfill
\begin{subfigure}{0.5\textwidth}
\centering
    \includegraphics[scale=0.21, trim={3.5cm 0.0cm 0.0cm 0.0cm},clip]{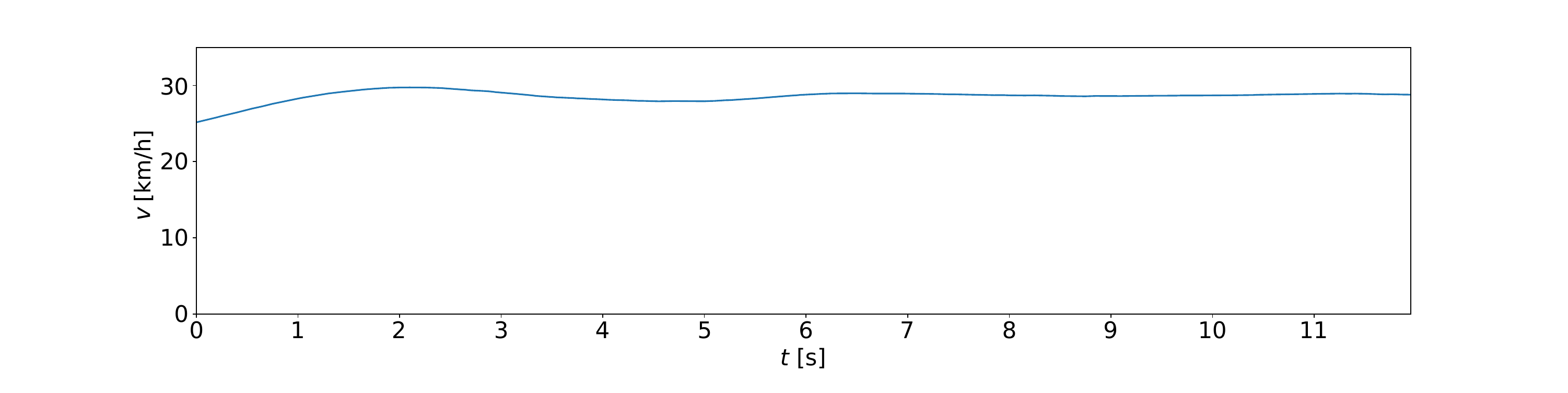}
    \caption*{Speed}
\end{subfigure}
\hfill
\begin{subfigure}{0.5\textwidth}
\centering
    \includegraphics[scale=0.21, trim={3.5cm 0.0cm 0.0cm 0.0cm},clip]{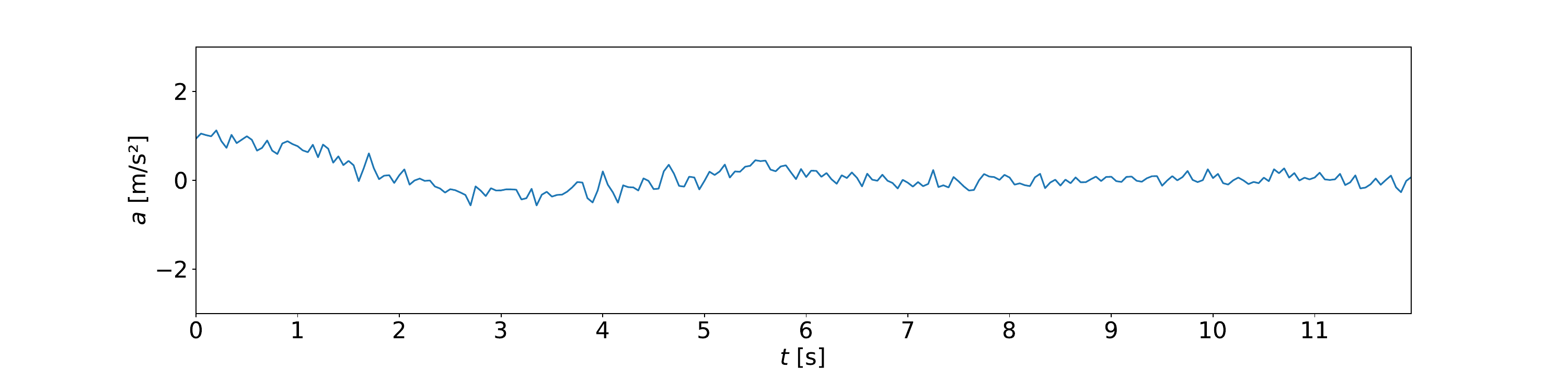}
    \caption*{Acceleration}
\end{subfigure}
\hfill
\begin{subfigure}{0.5\textwidth}
\centering
    \includegraphics[scale=0.21, trim={3.5cm 0.0cm 0.0cm 0.0cm},clip]{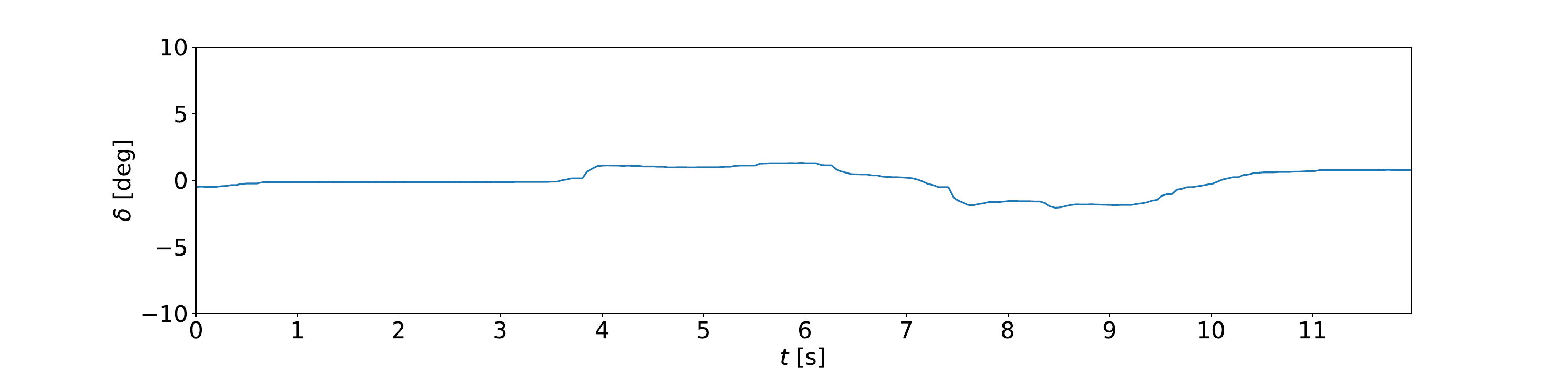}
    \caption*{Steering}
\end{subfigure}
\hfill
\caption{Object avoidance scenario.}
\label{fig:avoid}
\end{figure}

\begin{figure}[!h]
\centering
\begin{subfigure}{0.5\textwidth}
\centering
    \includegraphics[scale=0.24, trim={0.0cm 0.0cm 0.0cm 0.0cm},clip]{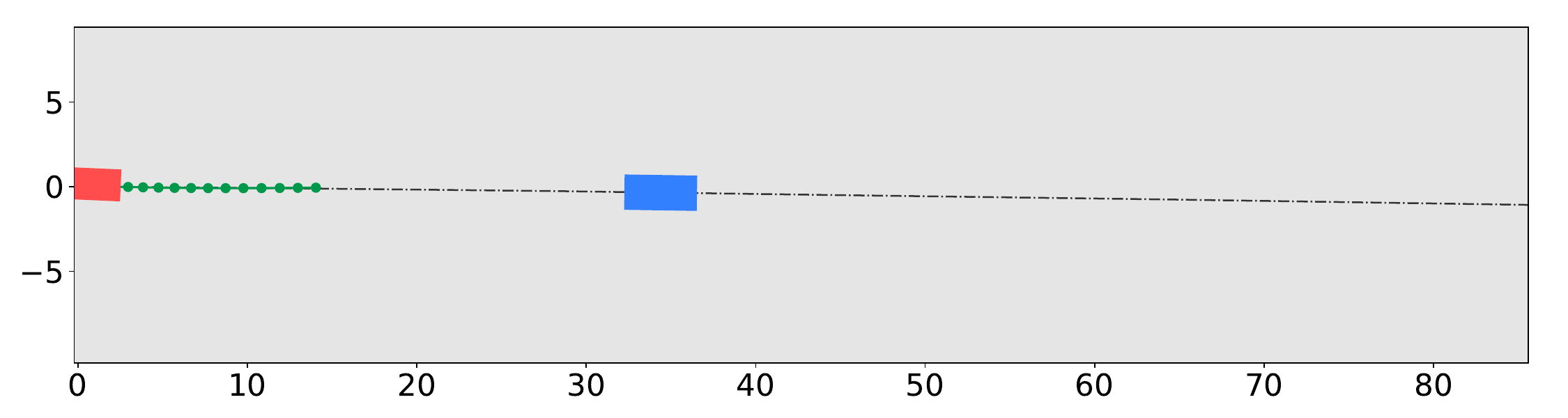}
    \caption*{$t\approx1$s}
\end{subfigure}
\hfill
\begin{subfigure}{0.5\textwidth}
\centering
    \includegraphics[scale=0.24, trim={0.0cm 0.0cm 0.0cm 0.0cm},clip]{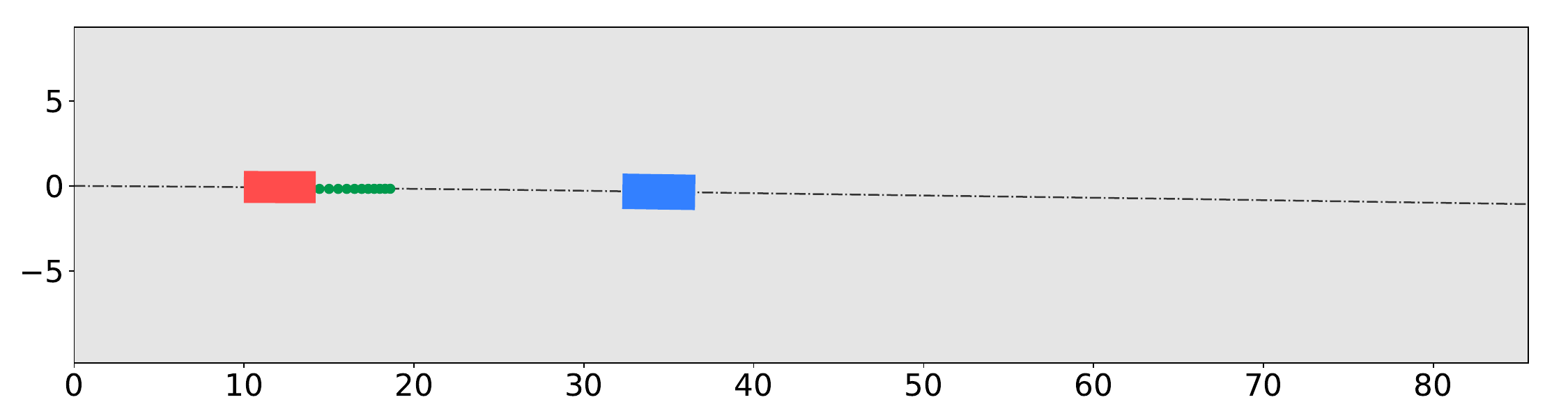}
    \caption*{$t\approx5$s}
\end{subfigure}
\hfill
\begin{subfigure}{0.5\textwidth}
\centering
    \includegraphics[scale=0.24, trim={0.0cm 0.0cm 0.0cm 0.0cm},clip]{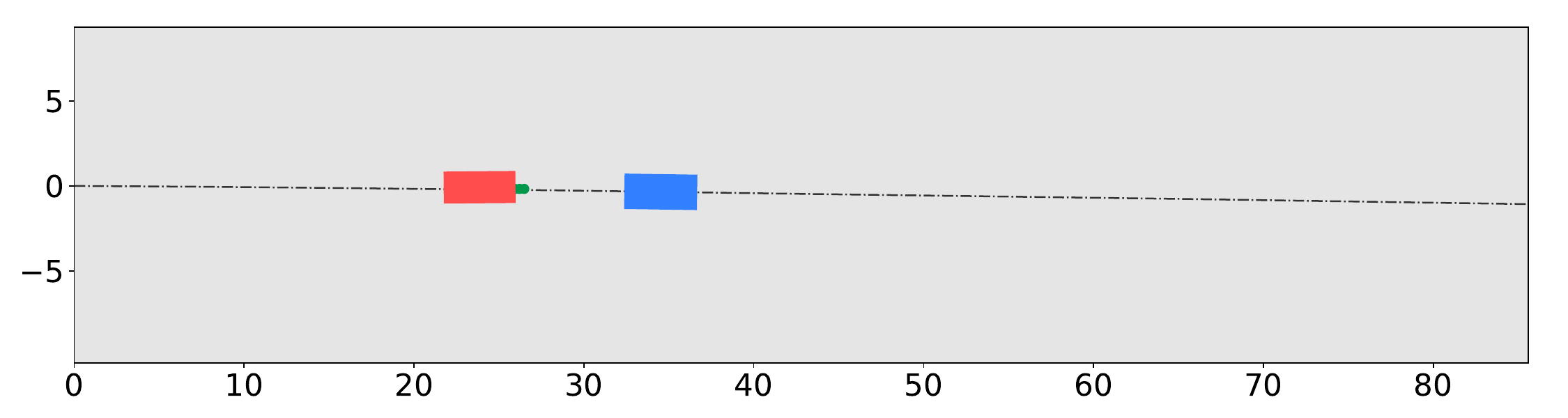}
    \caption*{$t\approx18$s}
\end{subfigure}
\hfill
\begin{subfigure}{0.5\textwidth}
\centering
    \includegraphics[scale=0.24, trim={0.0cm 0.0cm 0.0cm 0.0cm},clip]{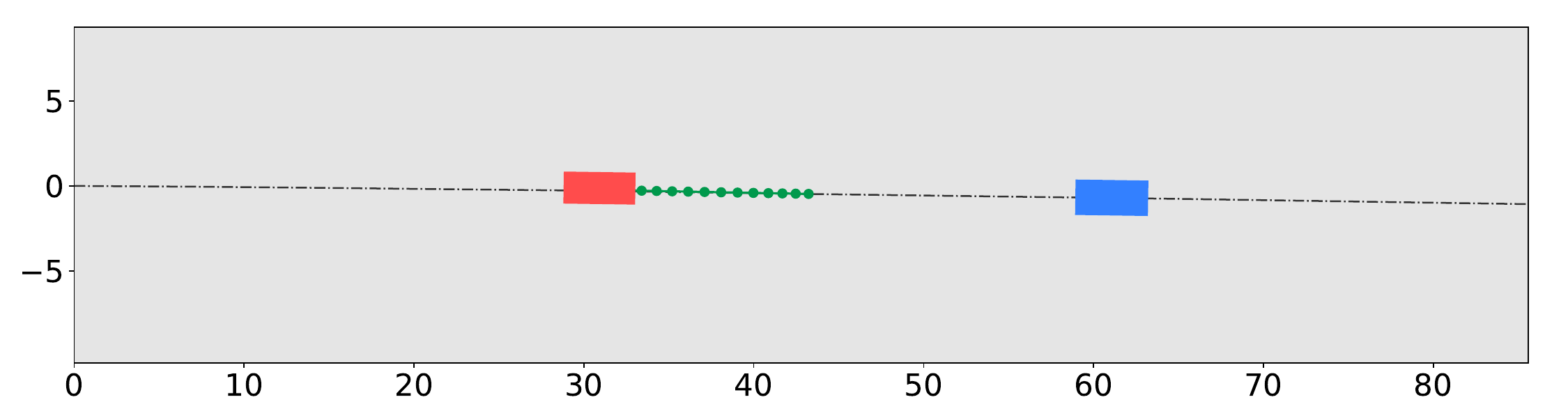}
    \caption*{$t\approx28$s}
\end{subfigure}
\hfill
\begin{subfigure}{0.5\textwidth}
\centering
    \includegraphics[scale=0.21, trim={3.5cm 0.0cm 0.0cm 0.0cm},clip]{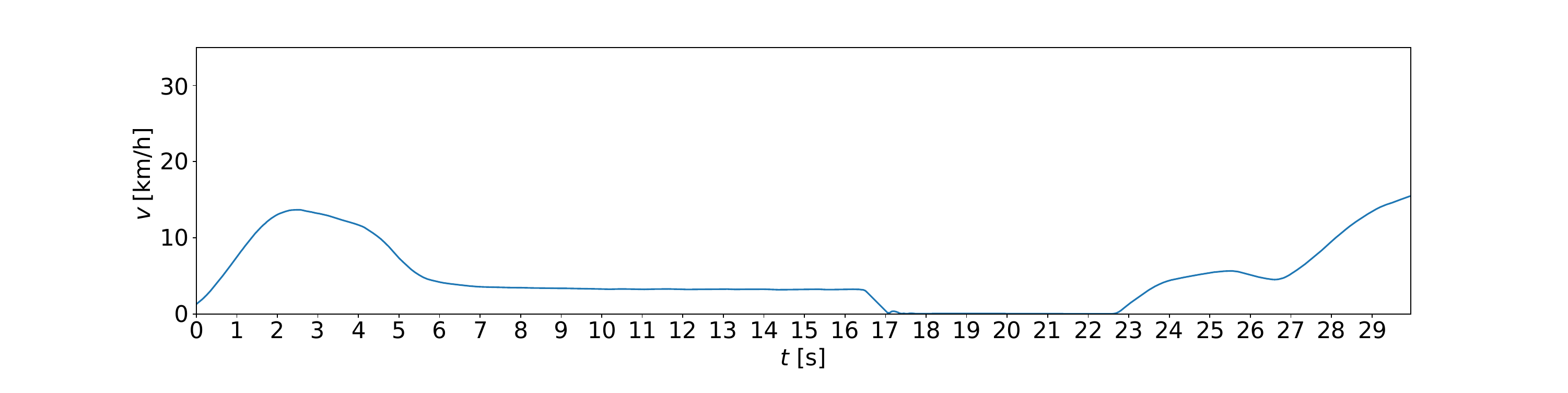}
    \caption*{Speed}
\end{subfigure}
\hfill
\begin{subfigure}{0.5\textwidth}
\centering
    \includegraphics[scale=0.21, trim={3.5cm 0.0cm 0.0cm 0.0cm},clip]{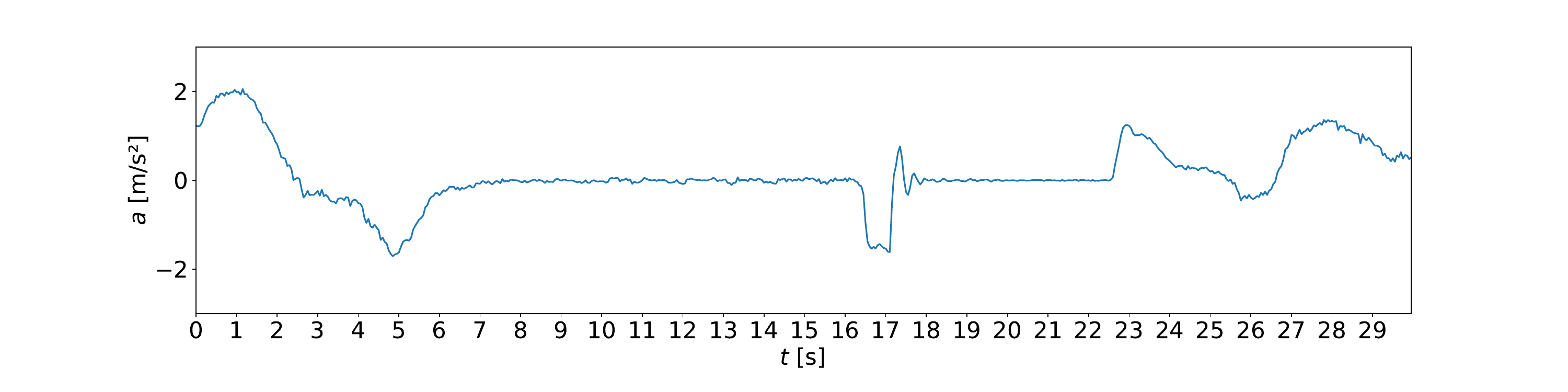}
    \caption*{Acceleration}
\end{subfigure}
\hfill
\begin{subfigure}{0.5\textwidth}
\centering
    \includegraphics[scale=0.21, trim={3.5cm 0.0cm 0.0cm 0.0cm},clip]{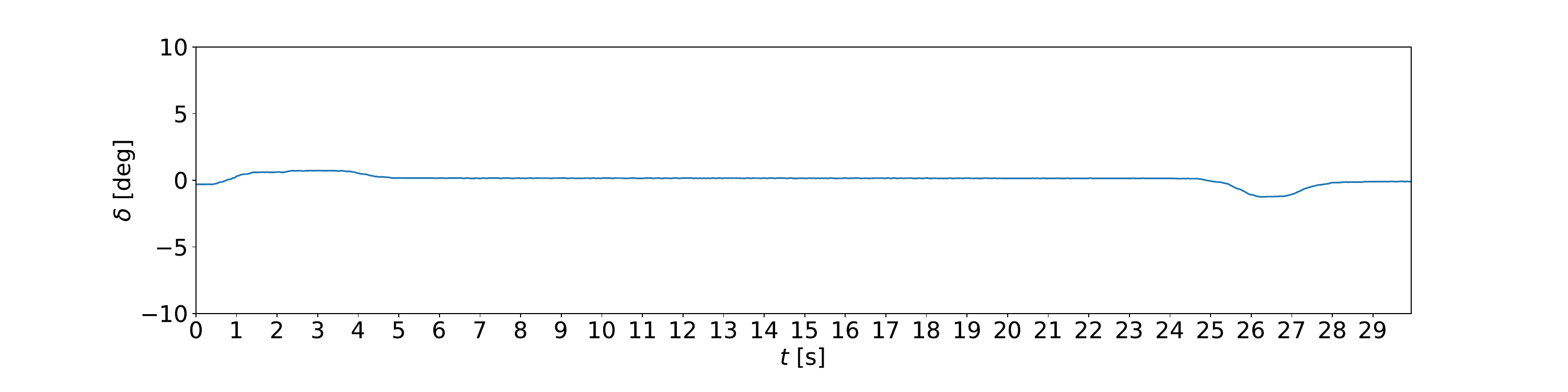}
    \caption*{Steering}
\end{subfigure}
\hfill
\caption{Object following scenario.}
\label{fig:follow}
\end{figure}

For each scenario, the generated trajectory and the vehicle state were recorded along the path. Snapshots of the results are depicted on Figures \ref{fig:merge}, \ref{fig:avoid} and \ref{fig:follow}. The path is represented by the dotted line, the ego vehicle and the virtual object are represented by a red and blue box respectively. The green line represents the generated MPPI trajectory at the given vehicle position. In addition, the realised steering, speed and acceleration are provided for each scenario as to better portray the actual behaviour of the vehicle over the course of the tests. The produced trajectories are smooth and they respect the prescribed acceleration and speed as this was guaranteed by this MPPI implementation. As it can be seen on the realised speed of Figures \ref{fig:merge}, \ref{fig:avoid} and \ref{fig:follow}, the vehicle converged toward the target speed of 30km/h and never exceeded it. The realised steering also shows a smooth operation for all three scenarios and never exceeded 10 degrees. The trajectories are proven safe as they not only avoid objects but also maintain a safe distance, and no collision occurred during the tests. It is important to note, however, that while the constraints on acceleration and steering rate ensure that the generated trajectory remains fairly easy to follow, and thus feasible, this cannot be guaranteed as the bicycle model used constitutes an estimation of the vehicle dynamics. The scope of this paper was to use the MPPI as a motion planner and therefore another module was in charge of the control of the vehicle. This module also has imperfections which could translate into a realised trajectory deviating from the target one. This happened during the tests and by looking at the realised acceleration on Figures \ref{fig:merge} and \ref{fig:follow}, it can be seen that the initial acceleration from a standstill exceeded the maximum value of $a_{max}$ by reaching almost 2. The minimum acceleration $a_{min}$, however, was never exceeded.

\section{Conclusion and Future Works}
This paper introduces a real-time and safe MPPI-based motion planner for autonomous driving. The method can handle obstacles and guarantees bounds for speed, acceleration and steering rate. Three primitive driving scenarios were proposed and tested on a real track at a maximum of speed of 30km/h. The results show that the method can perfectly handle those base scenarios and it achieved a smooth and accurate ride overall but also avoided collisions and kept a safe distance from obstacles. A slight deviation between the MPPI trajectories and the actual trajectories, especially for the acceleration, was observed and the underlying control module could be at fault. Future works should focus on more complex and realistic scenarios involving a more dynamic environment such as a lane change between moving vehicle. More work should also be done to better model the dynamics of the vehicle in the MPPI motion model.

\label{sec:conclusion}

\section*{Acknowledgment}
The authors would like to thank the Centre de Formation pour Conducteurs, Luxembourg for providing us access their training track. The authors would also like to thank Ihab S. Mohamed, Vehicle Autonomy and Intelligence Lab, Indiana University for the valuable discussion about MPPI.
\bibliographystyle{ieeetr}
\bibliography{bibliography} 

\end{document}